\documentclass[conference]{IEEEtran}
\IEEEoverridecommandlockouts

\usepackage{cite}
\usepackage{amsmath,amssymb,amsfonts}
\usepackage{algorithmic}
\usepackage{graphicx}
\usepackage{textcomp}
\usepackage{xcolor}
\usepackage{booktabs}
\usepackage{array}
\usepackage{url}
\usepackage{colortbl}

\def\BibTeX{{\rm B\kern-.05em{\sc i\kern-.025em b}\kern-.08em
    T\kern-.1667em\lower.7ex\hbox{E}\kern-.125emX}}

\newcommand{\cellGood}[1]{\cellcolor{green!22}#1}
\newcommand{\cellMid}[1]{\cellcolor{yellow!35}#1}
\newcommand{\cellBad}[1]{\cellcolor{orange!45}#1}
\newcommand{\cellVBad}[1]{\cellcolor{red!45}#1}

\begin{document}

\title{SemVerBench: Benchmarking LLM Comprehension of\\Version-Constraint Resolution Semantics}

\author{\IEEEauthorblockN{Qibai Chen}
\IEEEauthorblockA{\textit{Independent Researcher}\\
United States\\
qibaic@alumni.cmu.edu}
\and
\IEEEauthorblockN{Zeming Liu}
\IEEEauthorblockA{\textit{Brown University}\\
United States\\
zeming\_liu@brown.edu}
\thanks{Accepted at the 38th IEEE International Conference on Tools with
Artificial Intelligence (ICTAI 2026). This is the authors' preprint version.}}

\maketitle
\thispagestyle{plain}
\pagestyle{plain}

\begin{abstract}
Large language model (LLM) coding agents resolve dependency versions
constantly---deciding whether an installed version satisfies a declared
constraint such as \texttt{\^{}1.2.3} or \texttt{>=2.0,<3}---yet their grasp of
version-constraint \emph{semantics} has never been measured directly. We
introduce \textbf{SemVerBench}, the first benchmark of LLM version-constraint
resolution semantics across three ecosystems (npm, PEP~440, Cargo). SemVerBench
contains 240 machine-checkable items with unique answers, built author-neutrally
from four balanced sources (each ecosystem's official test suite plus three
frontier LLM proposers), and labeled by a non-circular two-implementation oracle
rather than by humans. Evaluating a six-model panel, we find systematic,
predictable per-mechanism blind spots rather than diffuse noise: a
partial-comparator carry rule (\texttt{>1.2}~$\equiv$~\texttt{>=1.3.0}) traps
\emph{every} model on Cargo (all near 60\%), and although PEP~440 standard prefix
matching is universal, on zero-pad/post-release corner cases GPT-5.1 collapses
(0/26 zero-pad cases) while Claude stays at 97--100\% (verified on a 67-item
oracle-validated set). Opus significantly outperforms all other models, and Sonnet
outperforms the OpenAI models (McNemar). Crucially, on the more parsimonious of two
readings our design cannot fully separate, the failures reflect an
\emph{activation/application} gap rather than a knowledge gap:
injecting the relevant rule recovers most errors and a light correct hint corrects
models that already failed, whereas interval decomposition does not help, and
models are at ceiling on the basic forms of the same rules. An author-stratified
analysis finds no statistically significant self-favoritism (the largest own-item
advantage, GPT-4.1 $+10$ points, is n.s.; the only significant author effect is
Gemini scoring lower on its own items), and the blind spots hold on official-suite
and cross-family items.
Because the task is verifiable and a free, 100\%-correct resolver exists, tool
delegation reaches $\approx$100\%. The takeaway: coding agents should delegate
version resolution to a resolver, not reason about versions in-head.
\end{abstract}


\section{Introduction}
\label{sec:intro}

GPT-5.1, a frontier reasoning model, handles the standard PEP~440 prefix form
\texttt{==2.*} perfectly---and yet, asked the zero-pad corner case ``does
\texttt{2} satisfy \texttt{==2.0.*}?'' (PEP~440 says yes, because it zero-pads the
release segment), it is wrong on \emph{all 26} such cases we tested, while Claude
scores 97--100\%. A free, deterministic resolver answers every such query in
microseconds, and the same model that fails the corner case aces the basic form.
This is not an isolated glitch. Across three packaging ecosystems and six frontier
models, we find that LLMs
\emph{systematically and predictably misapply version-constraint rules whose
simple forms they handle perfectly}.

LLM coding agents resolve dependency versions all the time: they read a lockfile,
judge whether an upgrade is admissible under \texttt{\^{}1.2.3}, pick a version
satisfying \texttt{>=2.0,<3}, or reason about whether a security fix falls inside a
declared range. These are high-frequency operations in every package manager (npm,
pip, Cargo, and beyond) and every coding assistant that touches dependencies. When
an agent gets version semantics wrong, the consequence is not a cosmetic typo but an
incorrect dependency decision---an unsafe version range, a missed patch, or a build
that silently resolves to the wrong release. Such errors are supply-chain-relevant:
empirically, a substantial fraction of dependent releases are broken by upstream
version/breaking-change issues~\cite{breakingchanges}. Despite this, the
\emph{semantics} of version-constraint resolution---as opposed to the upstream task
of \emph{inferring} which dependencies a repository needs---has never been
benchmarked.

We close that gap with \textbf{SemVerBench}. The task is deliberately narrow and
unambiguous: given a version $V$ and a constraint $C$ in ecosystem
$E\in\{\text{npm}, \text{PEP~440}, \text{Cargo}\}$, decide whether $V$ satisfies
$C$. Every item has a unique, machine-checkable answer. This narrowness is a
feature: it removes the confounds of open-ended generation and lets us attribute
every error to constraint semantics rather than to formatting, taste, or
ambiguity. The narrow surface has broad impact, because version resolution is
universal.

A first reaction might be that an LLM simply ``cannot compute'' such things. The
evidence points the other way, along three converging lines. First, models are at
or near ceiling on the basic forms of every grammar (plain membership, epochs,
exact pins), so they demonstrably command the underlying concepts; the failures
are concentrated in corner cases of rules whose simple forms they handle
perfectly. Second, a \emph{light, hedged} hint---``I think the answer
is\,\ldots'', not a forceful instruction---is enough to correct most
previously-failed items, which is more consistent with nudging a latent competence
than with teaching a missing one. Third, supplying the rule itself helps, whereas
supplying additional reasoning \emph{structure} (interval decomposition) does not:
the missing ingredient is the availability of the right rule in context, not more
reasoning steps. Together these are more consistent with an
\emph{activation/application} gap than with a capability or knowledge gap---a reading that aligns with work on
model self-knowledge~\cite{lmknow}, though, as we discuss in
Section~\ref{sec:limits}, our design cannot fully separate activating latent
knowledge from supplying missing knowledge.

Our findings are also structured, not diffuse. Difficulty concentrates in
specific mechanisms: basic membership, epochs, and plain ranges are at or near
ceiling for all models, while a handful of mechanisms account for nearly all
errors. Two stand out. First, a \emph{universal cross-ecosystem trap}: a partial
comparator rounds up, so \texttt{>1.2} means \texttt{>=1.3.0} and \texttt{>1}
means \texttt{>=2.0.0}; on Cargo, \emph{every} model---including the strongest---
falls to roughly 60\% on this mechanism. Second, a \emph{vendor-specific blind
spot}: PEP~440 standard prefix matching (\texttt{==N.*}) is universal---all six
models handle it---but its corner cases (zero-pad versions like \texttt{2} against
\texttt{==2.0.*}, and post-releases) form a sharp vendor-specific blind spot that
GPT-5.1 fails systematically while Claude does not. These are not random; they are
reproducible signatures of how each model family internalizes (or fails to
internalize) a documented rule.

\noindent\textbf{Contributions.}
\begin{enumerate}
  \item \textbf{SemVerBench}, the first benchmark of LLM
  version-constraint / semver resolution \emph{semantics}, spanning npm,
  PEP~440, and Cargo: 240 machine-checkable items with unique answers,
  constructed author-neutrally and labeled by a non-circular two-implementation
  oracle.
  \item A characterization of \textbf{systematic, per-mechanism blind spots}
  across six frontier models, including a universal cross-ecosystem
  partial-comparator trap and a sharp, vendor-specific prefix-match failure, with
  statistical significance (McNemar) and an author-stratified analysis showing no
  significant item-proposer self-favoritism.
  \item An \textbf{actionable diagnosis}: the errors look like
  activation/application gaps, recoverable by rule-injection and corrected by
  light accurate hints, and
  eliminated by \emph{delegating to the resolver} (tool use reaches
  $\approx$100\%).
\end{enumerate}

\section{Related Work}

\noindent\textbf{LLMs on software-engineering tasks.} LLMs are increasingly
evaluated on real software-engineering work---SWE-bench~\cite{swebench} measures
whether they resolve actual GitHub issues end-to-end---but such benchmarks bundle
many sub-competences behind a single pass/fail outcome. SemVerBench instead
isolates one specific, high-frequency, verifiable sub-competence: deciding version
membership.

\noindent\textbf{Dependency benchmarks for LLMs.} Recent work evaluates LLMs on
\emph{dependency inference}---which packages a repository requires. DI-BENCH~\cite{dibench}
scores whether a generated manifest lets a repository build and pass its tests, a
coarse end-to-end signal that does not isolate whether any version constraint was
understood; DependEval~\cite{dependeval} probes dependency-graph
\emph{understanding} (which modules depend on which), not the semantics of a single
specifier. Both ask which dependencies are needed; neither probes whether a version
correctly resolves against a constraint---our downstream, orthogonal task. To our
knowledge no prior benchmark targets version/semver resolution semantics at this
granularity.

\noindent\textbf{Constraint-following and constraint-reasoning benchmarks.} A
parallel line studies whether LLMs \emph{follow} constraints. CFBench~\cite{cfbench}
covers natural-language instruction constraints (format, length, content) judged on
free-form text, not a formal predicate with a single truth value;
ConstraintBench~\cite{constraintbench} (Gurobi-verified, operations-research)
concerns search over a feasible region, not specifier grammar; and ACS~\cite{acs}
uses an LLM as a soft judge of constraint-satisfaction in open-ended answers---the
opposite of our deterministic oracle. SemVerBench instead targets a small, formally
specified, verifiable semantics with a unique answer per item, labeled by code.

\noindent\textbf{Premise sensitivity and sycophancy.} Our controlled
false-premise study (Section~\ref{sec:f3}) is adjacent to the sycophancy
literature~\cite{sycophancy}, which shows models capitulating to user opinions on
subjective prompts. Our setting differs: the task is objective and verifiable, so a
``hint'' is either correct or incorrect against ground truth. We find an
\emph{asymmetric} effect---models adopt correct hints strongly and resist weak
incorrect ones---consistent with the activation/application account rather than
indiscriminate sycophancy.

\section{SemVerBench}
\label{sec:bench}

\subsection{Task}
An item is a triple $(E, V, C)$: ecosystem, version, and constraint. The model
must decide membership---does $V$ satisfy $C$?, which we write $V \models C$---and
end its response with \texttt{ANSWER: true} or \texttt{ANSWER: false}. Every item is a Boolean
membership query: there is exactly one correct answer, computable by a
deterministic resolver, and no item asks for free-form text, ranking, or
generation. Scoring is exact match against the oracle label after extracting the
final verdict. Parse failures (no extractable verdict) are tallied separately
rather than scored wrong, so comprehension is not conflated with output formatting.
This minimal, verifiable framing lets us attribute every error to constraint
semantics: a wrong answer cannot be excused as taste, ambiguity, or an unlucky
phrasing, because a one-line resolver call settles the item.

\subsection{Ecosystems and Their Grammars}
SemVerBench covers three ecosystems whose constraint grammars overlap
superficially but differ in subtle, consequential ways. \textbf{npm} uses the
node-semver range grammar: a range is a disjunction of hyphen ranges and
comparator sets, with sugar for caret (\texttt{\^{}}), tilde (\texttt{\~{}}), and
\texttt{x}-ranges; the tricky parts are caret/tilde behavior when leading
components are zero, partial comparators, and an opt-in prerelease policy.
\textbf{Python PEP~440} defines version \emph{specifiers} over a structured
version (epoch, release, pre/post/dev segments) with operators including
\texttt{==}, \texttt{!=}, \texttt{\~{}=} (compatible release), and the
prefix-match form \texttt{==N.*}; its subtleties include epoch ordering,
prefix matching on the release segment, and the rule that ordered comparators
exclude adjacent pre/post releases of the boundary version. \textbf{Cargo}'s
\texttt{VersionReq} resembles npm semantically (default caret behavior) but its
own treatment of partial comparators is the dominant source of difficulty in our
data. These three grammars give a small but genuinely diverse semantic surface.

\subsection{The Two-Axis Taxonomy}
Each item is tagged along two orthogonal axes. The \emph{syntax} axis records the
surface grammatical form of the specifier---caret/tilde sugar, a bounded
comparator (e.g., \texttt{>=1.2,<2.0}), a wildcard/prefix form (e.g.,
\texttt{2.*}), or an exact pin---i.e., \emph{how the constraint is written}. The
\emph{mechanism} axis records the underlying semantic rule the item exercises---
e.g., the partial-comparator carry or ordered-comparison exclusion---i.e.,
\emph{what resolving the constraint requires the model to know}. The two are
orthogonal: the same syntactic form can exercise different mechanisms, and the same
mechanism can surface under different syntax. We organize the analysis by the
mechanism axis, since the mechanism isolates the semantic difficulty; the syntax
axis is recorded only for coverage and balance, as two syntactically identical
items may be trivial or hard depending on which rule they hinge on. Key mechanisms
include:

\begin{itemize}
  \item \textbf{Caret/tilde and magic-zero} (\texttt{\^{}}, \texttt{\~{}}):
  compatibility ranges whose semantics change when leading components are zero
  (e.g., \texttt{\^{}0.2.3}).
  \item \textbf{Partial-comparator carry}: a partial comparator rounds up, so
  \texttt{>1.2}~$\equiv$~\texttt{>=1.3.0} and \texttt{>1}~$\equiv$~\texttt{>=2.0.0}.
  \item \textbf{Prerelease admission}: when a prerelease such as
  \texttt{1.2.0-rc.1} is or is not admitted by a range.
  \item \textbf{PEP~440 ordered-comparison exclusion}: \texttt{>V} does not match
  a post-release of $V$, and \texttt{<V} does not match a pre-release of $V$.
  \item \textbf{Epoch}: PEP~440 epoch-prefixed versions (e.g., \texttt{1!1.0}).
  \item \textbf{Prefix match}: PEP~440 \texttt{==N.*} / \texttt{!=N.*} matching on
  the release-segment prefix.
\end{itemize}

\subsection{Author-Neutral Construction}
SemVerBench has \textbf{240 items, 80 per ecosystem}, drawn from \textbf{four
author-neutral sources, 20 items each}. The first source is each ecosystem's
\emph{official test suite}---the regression tests shipped with the canonical
resolver implementation. The other three sources are frontier LLM
\emph{proposers} (Claude, GPT, Gemini), each given the \emph{identical} prompt:
propose challenging version/constraint membership items within fixed per-ecosystem
topic areas (e.g., prerelease admission, caret/tilde ranges), optimizing for
difficulty. A proposer's suggested answer is \emph{discarded}; every item is
independently labeled by the oracle, so no model labels its own items. We call the
construction \emph{author-neutral} in that all three proposers receive the same
prompt and topics---no vendor is favored---and any residual selection bias is
measured post-hoc (Section~\ref{sec:coi}); we do not claim the prompt is
content-free. This four-way design diversifies difficulty (official
suites anchor documented behavior; LLM proposers surface corner cases) and, since
every item's proposer is recorded, enables the conflict-of-interest analysis in
Section~\ref{sec:coi}. Beyond these main 240, we additionally build a 67-item
author-constructed, oracle-validated prefix-match set used solely for the focused
validation in Finding~3 (detailed there); it is not part of the main accuracy
numbers.

\subsection{Ground Truth: A Non-Circular Two-Implementation Oracle}
We use \emph{no human labels}. Each item's answer is computed by \emph{two
independent reference implementations} per ecosystem and retained only when they
agree:
\begin{itemize}
  \item \textbf{npm}: \texttt{node-semver} $\times$ \texttt{semantic\_version}.
  \item \textbf{PEP~440}: \texttt{packaging} $\times$ \texttt{pep440\_rs}.
  \item \textbf{Cargo}: the Rust \texttt{semver} crate $\times$
  \texttt{semantic\_version}.
\end{itemize}
The oracle is non-circular in two senses. First, the two implementations per
ecosystem are independently authored projects in different languages---for PEP~440,
\texttt{packaging} is the reference Python implementation while \texttt{pep440\_rs}
is an independent Rust reimplementation---so an item is labeled only when two
codebases sharing no implementation lineage agree, which filters out single-library
quirks and bugs. Second, the oracle is independent of the \emph{subjects}: no panel
model labels any item, and three of the four item sources are the panel models only
as \emph{proposers}, never as graders. We additionally
apply, for PEP~440, a \emph{prerelease-policy-robustness filter}: items whose
answer would depend on a resolver's prerelease-inclusion \emph{policy} (rather
than on the specifier semantics themselves) are dropped, so that every retained
PEP~440 item has a policy-independent answer and cannot be ``wrong'' merely
because of a defensible policy choice.

\subsection{Evaluation Protocol}
We evaluate a \textbf{six-model panel}: Claude-Opus, Claude-Sonnet, GPT-5.1,
GPT-4.1, GPT-4o, and Gemini-2.5-Pro. For reproducibility, the exact identifiers
are \texttt{claude-opus-4-6}, \texttt{claude-sonnet-4-6}, \texttt{gpt-5.1},
\texttt{gpt-4o}, \texttt{gpt-4.1}, and \texttt{gemini-2.5-pro}, all evaluated in
June 2026. The oracle libraries are likewise pinned---\texttt{node-semver} 7.8.1,
\texttt{packaging} 26.2, \texttt{pep440\_rs} 0.6.5, the Rust \texttt{semver} crate
1.0.28, and \texttt{semantic\_version} 2.10.0---because partial-comparator and
prerelease behavior can change across library releases, so our labels are only as
stable as these pins. Each item is posed as a single question---whether version
$V$ satisfies constraint $C$---with the model instructed to end its answer with
\texttt{ANSWER: true} or \texttt{ANSWER: false}; the prompt-sensitivity ablation
(Section~\ref{sec:limits}, accuracy shifts $\leq$4.2 points across three phrasings)
confirms the results are not an artifact of prompt phrasing. All models are queried
at temperature 0
with \textbf{three independent runs} over the full 240 items. At temperature 0 the
three runs are nearly identical (run-to-run spread 0.4--3.3 points across models),
so across-run variance understates true uncertainty; we therefore report
\textbf{Wilson 95\% confidence intervals} that treat each model's 3-run mean
accuracy as a binomial proportion over the $n=240$ items, reflecting benchmark
sampling uncertainty. For paired model comparisons we use the \textbf{McNemar
exact-binomial test}, computed on per-item correctness after taking the majority
vote across the three runs ($n=240$ paired items). There were \textbf{0 API
errors} and a parse-failure rate of
\textbf{0.30\%}. The 240 items span \textbf{9 mechanisms} and \textbf{18 syntax
tags}, and are mildly imbalanced toward ``false'': 103 true (42.9\%) and 137
false (57.1\%). A majority-class predictor that always answers ``false'' therefore
scores \textbf{57.1\%} (per-ecosystem false rates: npm 52.5\%, PEP~440 58.8\%,
Cargo 60.0\%); this is the chance floor against which all accuracies below should
be read.

\section{Findings}
\label{sec:findings}

Table~\ref{tab:main} reports overall accuracy. Claude-Opus leads at 90.6
[86.0, 93.5], followed by Claude-Sonnet at 86.8 [81.8, 90.4]; the OpenAI models
and Gemini cluster between 80\% and 83\%. By McNemar, Opus is significantly better
than every other model; Sonnet is significantly better than each OpenAI model;
Sonnet versus Gemini is not significant; and the OpenAI models and Gemini are
mutually not significant. Our significance claims rest on the paired McNemar test
(per-item majority vote), which controls for item difficulty and can detect
differences that overlapping marginal CIs mask.

\begin{table}[t]
\centering
\caption{Overall accuracy with Wilson 95\% CIs treating the 3-run mean accuracy
as a binomial proportion over $n=240$ items. McNemar: Opus $>$ all ($p<0.01$--$.001$);
Sonnet $>$ each OpenAI model ($p<0.05$--$.01$); Sonnet vs.\ Gemini n.s.; OpenAI \&
Gemini mutually n.s.}
\label{tab:main}
\begin{tabular}{lcc}
\toprule
\textbf{Model} & \textbf{Acc.\ (\%)} & \textbf{95\% CI} \\
\midrule
Claude-Opus    & 90.6 & [86.0, 93.5] \\
Claude-Sonnet  & 86.8 & [81.8, 90.4] \\
Gemini-2.5-Pro & 82.9 & [77.6, 87.2] \\
GPT-4.1        & 82.2 & [76.7, 86.4] \\
GPT-5.1        & 80.4 & [74.9, 84.9] \\
GPT-4o         & 80.1 & [74.5, 84.6] \\
\bottomrule
\end{tabular}
\end{table}

These overall accuracies of 80--91\% sit well above the 57.1\% majority-class
floor, confirming genuine competence on the bulk of items: the benchmark is not
solved by exploiting label imbalance. But this is only the aggregate picture: as
Findings~2--4 show, several blind-spot buckets do \emph{not} clear the
majority-class floor at all.

Indeed, aggregate accuracy hides the structure that is the point of this paper.
Table~\ref{tab:heatmap} breaks accuracy down per (ecosystem $\times$ mechanism).
Plain membership,
epochs, and basic ranges sit at or near 100\% for all models; the difficulty is
\emph{concentrated} in a few mechanisms. Crucially, the headline blind-spot
buckets fall close to their per-bucket chance floors (the \textbf{maj\%} column of
Table~\ref{tab:heatmap}): Cargo partial-carry at $\approx$60\% sits only about 4
points above that bucket's 56\% majority-class rate, so the models retain almost no
signal beyond guessing (Sonnet's 59\% is essentially at the floor). npm
partial-carry corroborates the mechanism: though its 80\% clears its 60\%
majority-class rate, all six models are pinned to the identical 80\%, so the
difficulty is a property of the carry rule, not of any one model. The PEP~440
ordered-exclusion bucket (60--70\%) is similarly near chance for weaker models. We
discuss each in turn.

\begin{table*}[t]
\centering
\caption{Per-(ecosystem $\times$ mechanism) accuracy (\%, pooled over 3 runs).
$n$ is the number of items in the bucket (summing to 240). \textbf{maj\%} is the
per-bucket majority-class rate (the score of always predicting that bucket's more
common label), i.e., the chance floor for that bucket. Bold rows are the
headline blind spots. Cell shading: green $\geq$90, yellow 75--89, orange 50--74,
red $<$50.}
\label{tab:heatmap}
\renewcommand{\arraystretch}{1.15}
\begin{tabular}{l ccc ccc r r}
\toprule
\textbf{ecosystem.mechanism} & \textbf{Opus} & \textbf{Sonnet} & \textbf{GPT-5.1} & \textbf{GPT-4.1} & \textbf{GPT-4o} & \textbf{Gemini} & \textbf{$n$} & \textbf{maj\%} \\
\midrule
cargo.bare\_caret                 & \cellGood{100} & \cellGood{100} & \cellGood{94}  & \cellGood{100} & \cellGood{100} & \cellMid{83}   & 6  & 50 \\
cargo.magic\_zero                 & \cellGood{100} & \cellGood{100} & \cellGood{93}  & \cellGood{94}  & \cellMid{87}   & \cellGood{94}  & 18 & 50 \\
\textbf{cargo.partial\_carry}     & \cellBad{63}   & \cellBad{59}   & \cellBad{62}   & \cellBad{63}   & \cellBad{63}   & \cellBad{63}   & 27 & 56 \\
cargo.plain                       & \cellGood{100} & \cellGood{100} & \cellGood{92}  & \cellGood{100} & \cellGood{97}  & \cellGood{92}  & 13 & 62 \\
cargo.prerelease                  & \cellMid{85}   & \cellGood{94}  & \cellMid{75}   & \cellMid{88}   & \cellMid{81}   & \cellGood{96}  & 16 & 81 \\
npm.magic\_zero                   & \cellGood{100} & \cellGood{96}  & \cellGood{96}  & \cellGood{95}  & \cellMid{81}   & \cellGood{98}  & 19 & 58 \\
npm.partial\_carry                & \cellMid{80}   & \cellMid{80}   & \cellMid{80}   & \cellMid{80}   & \cellMid{80}   & \cellMid{80}   & 20 & 60 \\
npm.plain                         & \cellGood{100} & \cellGood{100} & \cellGood{100} & \cellGood{100} & \cellGood{100} & \cellGood{100} & 9  & 56 \\
npm.prerelease                    & \cellGood{100} & \cellGood{98}  & \cellMid{88}   & \cellMid{86}   & \cellMid{85}   & \cellMid{86}   & 32 & 59 \\
pep440.compatible                 & \cellGood{100} & \cellMid{89}   & \cellMid{75}   & \cellMid{86}   & \cellMid{83}   & \cellGood{92}  & 12 & 58 \\
pep440.epoch                      & \cellGood{100} & \cellGood{100} & \cellGood{90}  & \cellGood{95}  & \cellGood{97}  & \cellGood{98}  & 20 & 65 \\
\textbf{pep440.ordered\_exclusion}& \cellMid{84}   & \cellBad{65}   & \cellBad{70}   & \cellBad{63}   & \cellBad{62}   & \cellBad{60}   & 35 & 91 \\
pep440.plain                      & \cellGood{94}  & \cellGood{100} & \cellGood{100} & \cellGood{94}  & \cellGood{94}  & \cellGood{100} & 6  & 67 \\
\textbf{pep440.prefix\_match}     & \cellGood{95}  & \cellGood{90}  & \cellVBad{19}  & \cellVBad{38}  & \cellBad{52}   & \cellBad{52}   & 7  & 86 \\
\bottomrule
\end{tabular}
\end{table*}

\subsection{Finding 1: A cross-vendor gap}
Claude models (Opus, Sonnet) significantly outperform the OpenAI and Google
models by paired McNemar testing (Opus $>$ all others $p<0.01$--$.001$; Sonnet $>$
each OpenAI model $p<0.05$--$.01$), so the gap is not a confidence-interval
artifact. Notably, competence here does not track model recency: the newest
OpenAI model, GPT-5.1 (80.4\%), ranks \emph{numerically} below GPT-4.1 (82.2\%),
though this within-vendor gap is \emph{not} statistically significant (the OpenAI
models and Gemini are mutually n.s.\ by McNemar, with overlapping Wilson CIs), so
we read it only as evidence that recency does not predict competence, not as a
reliable ordering. We cannot attribute the cross-vendor gap to specific training
differences---the models are closed---and need not: the point is \emph{not}
``Claude wins'' but that the spread itself is unjustified, since the task is
machine-checkable and a free, 100\%-correct resolver exists, so any in-head error
rate---for any vendor---is avoidable.

\subsection{Finding 2: Partial-comparator carry is a universal trap}
A partial comparator rounds up: \texttt{>1.2} is equivalent to \texttt{>=1.3.0},
and \texttt{>1} is equivalent to \texttt{>=2.0.0}. On npm, all six models land at
\emph{exactly} 80\% on this mechanism (16 of 20 items correct under 3-run
majority, for every model).
That cross-model uniformity is itself informative: when six independently trained
systems converge on the identical score, the difficulty is a property of the
\emph{mechanism}, not of any one model. On Cargo the effect is more severe:
\emph{every} model falls to roughly 60\%---Opus 63\%, Sonnet 59\%, and the
remaining four in the 62--63\% band. This is the clearest cross-ecosystem,
cross-vendor blind spot in the benchmark, and notably it does not spare the
strongest model. A representative failure: \texttt{1.0.5} $\models$ \texttt{>1} is
\emph{False} (because \texttt{>1} carries to \texttt{>=2.0.0}), yet models answer
\emph{True}, apparently treating \texttt{>1} as \texttt{>1.0.0}. The rule is short
and documented; the models simply do not apply it.

\subsection{Finding 3: Prefix-match corner cases are a sharp, vendor-specific blind spot}
PEP~440 prefix-matching corner cases produce the single most dramatic failure in
SemVerBench. On the main benchmark's prefix-match bucket ($n=7$ items), only
Claude handles it (Opus 95\%, Sonnet 90\%); everyone else drops sharply---GPT-5.1
to \textbf{19\%}, GPT-4.1 to 38\%, GPT-4o and Gemini to 52\%. These 7 items are
predominantly corner-type (zero-pad, post-release, multi-segment)---consistent with
the 21\% GPT-5.1 scores on the focused corner set below---while the standard form is
at 100\% for all models.

\noindent\textbf{Focused validation.} The main prefix-match bucket is small
($n=7$) yet carries the paper's most dramatic claim, so it warrants targeted
scrutiny (the other small buckets, e.g.\ \texttt{cargo.bare\_caret} $n=6$, are
near-ceiling and do not hinge on a few items). We therefore built a 67-item
prefix-matching validation set, separate from the main 240. The items are \emph{author-constructed}:
we programmatically enumerated and hand-designed $(\text{version}, \text{constraint})$
pairs covering two regimes---\emph{standard} prefix (the version has at least the
prefix's release segments and no post-release) and \emph{corner} cases (zero-pad,
post-release, multi-segment)---deliberately not drawn from the main benchmark's LLM
proposers, since a targeted validation is meant to probe specific sub-cases.
Labeling uses the identical oracle (\texttt{packaging} under three prerelease
policies, kept only if robust across all three, $\times$ \texttt{pep440\_rs}
consensus). The set comprises 34 standard $+$ 33 corner $=67$ items; the 33 corner
items are 26 zero-pad/post-release ``true'' cases (e.g.,
\texttt{2}~$\models$~\texttt{==2.0.*}, \texttt{2.0.post1}~$\models$~\texttt{==2.0.*})
plus 7 boundary ``false'' cases.

The blind spot is not a small-sample artifact (Table~\ref{tab:prefix}). On the 34
standard items \emph{all six models score 100\%}: every model handles the ordinary
\texttt{==2.*} form. The blind spot lives entirely in the corner cases, where it is
sharp and vendor-specific: GPT-5.1 scores \textbf{21\%}---and \textbf{0/26} on the
zero-pad/post-release ``true'' subset specifically---while Claude stays at
97--100\% and GPT-4o/4.1/Gemini sit in a 58--67\% middle band. The diagnosis is
concrete: GPT-5.1 systematically treats a version with fewer release segments than
the prefix (\texttt{2} vs.\ \texttt{==2.0.*}, which PEP~440 zero-pads) and
post-release versions as non-matching, judging every such case False. Because the
rule is well-specified, this is a vendor-specific internalization gap, not a
hardness property of the mechanism: one vendor has clearly learned the corner
cases and the others have not.

\begin{table}[t]
\centering
\caption{Focused prefix-match validation (\%, 3-run majority): 67 author-constructed,
oracle-validated PEP~440 prefix items, separate from the main 240
($34$ standard $+ 33$ corner). The last column is the $26$ zero-pad/post-release
``true'' cases that form a \emph{subset} of the $33$ corner items (the other $7$
corner items are boundary ``false'' cases).}
\label{tab:prefix}
\begin{tabular}{lccc}
\toprule
\textbf{Model} & \textbf{Standard} & \textbf{Corner} & \textbf{Zero-pad/post} \\
 & \textbf{($n{=}34$)} & \textbf{($n{=}33$)} & \textbf{(true, $\subset$33, $n{=}26$)} \\
\midrule
Claude-Opus    & 100 & 100 & 26/26 \\
Claude-Sonnet  & 100 & 97  & 25/26 \\
GPT-5.1        & 100 & 21  & 0/26  \\
GPT-4o         & 100 & 67  & 15/26 \\
GPT-4.1        & 100 & 58  & 12/26 \\
Gemini-2.5-Pro & 100 & 58  & 13/26 \\
\bottomrule
\end{tabular}
\end{table}

\subsection{Finding 4: Ordered-comparison exclusion cracks Sonnet}
The PEP~440 ordered-comparison exclusion rule states that \texttt{>V} does not
match a post-release of $V$ (e.g., \texttt{1.0.0.post1}), and \texttt{<V} does not
match a pre-release of $V$. This mechanism is broadly hard---most models land in
the 60--70\% range---and it specifically pulls Sonnet down from its 87\%-class
overall standing to \textbf{65\%}. Opus is more robust here at 84\%. A
representative failure: \texttt{1.0.0.post1} $\models$ \texttt{>1.0.0} is
\emph{False} under PEP~440, but models commonly answer \emph{True}.

\subsection{Finding 5: Conflict of interest: no significant self-favoritism}
\label{sec:coi}
Because three of our four item sources are themselves LLMs, a natural concern is
self-favoritism. Table~\ref{tab:coi} stratifies each model's accuracy by item
source---items proposed by its own vendor family ($n=60$), items proposed by the
\emph{other} LLM families ($n=120$, official suites excluded), and the official
test suites. We test each model's own-vs-other-LLM gap with a two-proportion test.
\emph{No model shows a statistically significant own-item advantage.} The gap is
numerically larger for the GPT family (up to $+10$ points for GPT-4.1), but even
that is not significant (GPT-4.1 $+10$, two-proportion $p\approx0.10$; all other
positive gaps $p>0.27$); at only $n=60$ own items these effects are
indistinguishable from noise. The Claude family's gaps are small ($\sim$$+3$). The
\emph{only} significant author effect runs the \emph{opposite} way: Gemini scores
significantly lower on its own items (70.0 vs.\ 83.9, about 14 points lower,
$p=0.02$)---its own items are the hardest for it.

There is thus no statistically significant self-favoritism; if anything, the one
significant effect is anti-favoritism. This does not threaten the headline
findings, for three reasons. (a) The blind spots---partial-comparator carry,
prefix corner cases, and ordered-comparison exclusion---appear on the
official-suite and cross-family items for \emph{every} model, not only on
self-authored ones. (b) The leaderboard ordering is not author-driven (Opus's
own-item gap is small and n.s.). (c) One full source---the official suites---has
no panel-model stake, and the two-implementation oracle labels every item
independently of any model.

\begin{table}[t]
\centering
\caption{Conflict-of-interest check: accuracy (\%, 3-run pooled) stratified by
item source. ``own'' = items proposed by the model's own vendor family ($n=60$);
``other'' = items proposed by the other LLM families ($n=120$, official suites
\emph{excluded}); ``official'' = the official test-suite sources. No own-vs-other
gap is significant (two-proportion test); the largest, GPT-4.1 $+10$, has
$p\approx0.10$. The only significant author effect is Gemini scoring \emph{lower}
on its own items ($p=0.02$).}
\label{tab:coi}
\begin{tabular}{lccc}
\toprule
\textbf{Model} & \textbf{own-LLM} & \textbf{other-LLM} & \textbf{official-suite} \\
\midrule
Claude-Opus    & 90.0 & 86.9 & 98.3 \\
Claude-Sonnet  & 84.4 & 81.7 & 99.4 \\
GPT-5.1        & 79.4 & 74.7 & 92.8 \\
GPT-4.1        & 84.4 & 74.4 & 95.6 \\
GPT-4o         & 80.6 & 72.2 & 95.6 \\
Gemini-2.5-Pro & 70.0 & 83.9 & 93.9 \\
\bottomrule
\end{tabular}
\end{table}

\subsection{Error analysis: recurring failure modes}
The mistakes are not random; Table~\ref{tab:heatmap} and the worked examples point
to three recurring failure \emph{modes}, each a specific misreading of a rule.
\textbf{(M1) Partial-comparator under-carry:} a partial comparator such as
\texttt{>1} is treated as \texttt{>1.0.0} instead of the correct \texttt{>=2.0.0},
so a version like \texttt{1.0.5} is wrongly admitted; this is the dominant Cargo
error and the source of the universal $\approx$60\% bucket. \textbf{(M2) Prefix
under-match:} a version with fewer release segments than the prefix (\texttt{2} vs.\
\texttt{==2.0.*}, which PEP~440 zero-pads) or a post-release (\texttt{2.0.post1}
vs.\ \texttt{==2.0.*}) is wrongly judged non-matching; this drives the GPT-5.1
prefix collapse (0/26 on these cases in the focused validation), while the
standard prefix form is handled correctly by all models. \textbf{(M3) Boundary
pre/post admission:} an ordered comparator is applied without the
adjacent-release exclusion, so \texttt{1.0.0.post1} is wrongly judged to satisfy
\texttt{>1.0.0} (and symmetrically a pre-release is wrongly admitted by \texttt{<V});
this mode underlies the ordered-exclusion bucket that cracks Sonnet to 65\%. All
three modes are corrected by rule-injection (Section~\ref{sec:mitig}).

By ecosystem, PEP~440 is hardest in aggregate---it concentrates M2 and M3 and has
the most mechanisms, so its blind-spot buckets dominate the lower half of
Table~\ref{tab:heatmap}. Cargo is easiest \emph{except} for the partial-carry
trap, which makes it a clean isolation of M1: its other mechanisms sit near
ceiling, leaving the carry rule as the lone signal. npm sits in between.

\section{Mitigation: Delegate to the Resolver}
\label{sec:mitig}

Having mapped the blind spots, we ask what fixes them. We compare four
conditions per model. \textbf{E0} is the in-head baseline---identical to the main
three-run accuracy in Table~\ref{tab:main}. \textbf{E1} (rule-injection) prepends
the relevant resolution rule to the prompt. \textbf{E2} (interval decomposition)
appends a three-step instruction: (1) rewrite the constraint as explicit version
intervals ($\geq,>,\leq,<$, or exact) under the ecosystem's rules, (2) write the
version's full precedence form, (3) check membership and answer. \textbf{E3} (tool-use)---the version-resolution instance
of the general idea that LLMs can be equipped to call external
tools~\cite{toolformer}---gives the model a \texttt{check\_constraint} tool that
wraps the official resolver. E1/E2/E3 are
single-run conditions. Results are in Table~\ref{tab:mitig}.

\begin{table}[t]
\centering
\caption{Mitigation accuracy (\%). E0 = in-head baseline (= Table~\ref{tab:main}
3-run mean). E1 = rule-injection, E2 = interval decomposition, E3 = tool-use
(official resolver). E1/E2/E3 are single-run.}
\label{tab:mitig}
\begin{tabular}{lcccc}
\toprule
\textbf{Model} & \textbf{E0} & \textbf{E1} & \textbf{E2} & \textbf{E3} \\
\midrule
Claude-Opus    & 90.6 & 100 & 89 & 100 \\
Claude-Sonnet  & 86.8 & 99  & 88 & 100 \\
GPT-5.1        & 80.4 & 91  & 79 & 100 \\
GPT-4.1        & 82.2 & 95  & 84 & 99  \\
GPT-4o         & 80.1 & 92  & 82 & 100 \\
Gemini-2.5-Pro & 82.9 & 95  & 85 & 100 \\
\bottomrule
\end{tabular}
\end{table}

Three results follow. \textbf{(i) E1 rule-injection produces a large lift} for
every model (e.g., GPT-5.1 $80.4\to91$, GPT-4.1 $82.2\to95$, Opus to 100\%):
making the relevant rule available in context recovers most errors. \textbf{(ii)
E2 interval decomposition produces essentially no lift} (several models move
within noise, some slightly down). The missing ingredient is therefore the right
rule in context, not additional reasoning steps; this is inconsistent with a pure
reasoning-steps deficit, though a single decomposition strategy cannot by itself
exclude every reasoning-based account. \textbf{(iii) E3 tool-use reaches
$\approx$100\%}: a \texttt{check\_constraint} tool wrapping the official resolver
makes the task trivially correct. In only 2 of 1440 tool-condition responses did
the model invoke the tool, receive the correct result, and still emit a wrong
verdict. The residual failure mode is thus not computation but ignoring or
misreading a correct tool output: even when the tool returned the right answer the
model occasionally failed to adopt it, so delegation works only if the agent
actually takes the tool's output as its verdict. When and why an agent invokes the
resolver at all---rather than answering from its weights---is a separate question
our fixed tool-use setup does not probe, and is worth future study.

\noindent\textbf{The diagnosis: an activation/application gap.} These mitigation
results complete the three converging lines of evidence laid out in
Section~\ref{sec:intro}: near-ceiling basic buckets, correction by a light hint,
and the rule (E1) helping while reasoning structure (E2) does not---together more
consistent with an activation/application gap than a capability or knowledge gap.
Two caveats apply. The light-hint lift may partly be a generic re-check trigger
rather than the hint's content (Section~\ref{sec:f3}); and because E1 and the TRUE
hint both place the rule in context, we cannot fully exclude
supplying-missing-knowledge (Section~\ref{sec:limits}). The activation reading is
the more parsimonious one, not a proof of prior possession.

The conclusion is direct: a free, deterministic, 100\%-correct resolver exists, so
any in-head error rate is unjustified and coding agents should delegate resolution
to that resolver rather than reason about versions themselves.

\noindent\textbf{Implications for coding-agent design.} The recipe follows from
Table~\ref{tab:mitig}. An agent that touches dependency versions should \emph{gate}
every membership or range decision through a resolver tool (the ecosystem's official
library)---the E3 condition, essentially free and exact---rather than emit a verdict
from its own weights; when no resolver is reachable (e.g., reasoning inline in
chat), the second-best mitigation is to inject the governing rule (E1). What it
should \emph{not} do is the current default of unscaffolded in-head reasoning (E0),
the regime in which the blind spots occur.

\subsection{Controlled false-premise sensitivity}
\label{sec:f3}
To test whether the activation/application account is sound and to probe robustness
to misinformation, we run a controlled false-premise study on a hard 61-item
subset. The subset is chosen by an explicit, pre-specified rule---the
lowest-baseline (E0) items per mechanism, i.e., the items models most often get
wrong at baseline---so that a premise has room to change the answer; it is not
hand-picked. Each item is presented with a weak, hedged hint of one of three kinds
plus a no-hint baseline: \textbf{TRUE} (a correct ``I think the answer
is\,\ldots'' hint), \textbf{NEUTRAL} (an uninformative hint), and \textbf{FALSE}
(an incorrect hedged hint). Because the subset is deliberately the hardest items,
its baseline accuracy is lower than the full-240 numbers in Table~\ref{tab:main};
it is a subset baseline and should not be conflated with Table~\ref{tab:main}.
Results are in Table~\ref{tab:f3}.

\begin{table}[t]
\centering
\caption{Controlled false-premise study (\%, hard 61-item subset, single run).
A weak hedged hint is TRUE, NEUTRAL, or FALSE. ``baseline'' is the no-hint
accuracy on this subset (lower than Table~\ref{tab:main}; do not conflate).}
\label{tab:f3}
\begin{tabular}{lcccc}
\toprule
\textbf{Model} & \textbf{baseline} & \textbf{TRUE} & \textbf{NEUTRAL} & \textbf{FALSE} \\
\midrule
Claude-Opus    & 80 & 93 & 75 & 80 \\
Claude-Sonnet  & 67 & 97 & 72 & 74 \\
GPT-5.1        & 39 & 56 & 49 & 57 \\
GPT-4.1        & 51 & 57 & 56 & 57 \\
GPT-4o         & 44 & 52 & 54 & 56 \\
Gemini-2.5-Pro & 57 & 66 & 57 & 62 \\
\bottomrule
\end{tabular}
\end{table}

The effect is \emph{asymmetric}. A TRUE hint corrects models massively---Sonnet
rises from 67\% to 97\% (roughly 18 of 20 previously-failed items corrected)---
whereas a weak FALSE hint barely misleads them (typically 0--2 flips, and FALSE
accuracy is often at or above baseline). For the weakest-baseline models, the mere
presence of any hedged premise---even an incorrect one---appears to trigger more
careful re-checking that a weak false premise is not assertive enough to override;
so part of their lift may be driven by the presence of a hint rather than its
content, an observation worth dedicated future study. Models thus \emph{adopt}
correct hints and \emph{resist} weak incorrect ones: we do not claim they are
``easily fooled by false premises.'' That a \emph{light} correct hint corrects so
many items fits the activation/application reading, and the resistance to weak
false hints shows robustness to mild misinformation when the answer is checkable.

\section{Limitations and Conclusion}
\label{sec:limits}

\noindent\textbf{Limitations.} First, the LLMs under test are also three of the
four item-proposer sources, so they are simultaneously subjects and authors. The
author-stratified analysis (Section~\ref{sec:coi}) finds no statistically
significant self-favoritism (the largest own-item gap, GPT-4.1 $+10$, is n.s.; the
only significant author effect is Gemini scoring \emph{lower} on its own items),
and we further mitigate with an official-suite source authored by no panel model
and a non-circular two-implementation oracle that labels every item independently.
The blind spots hold on official-suite and cross-family items, so they are not a
self-favoritism artifact. Second,
Cargo is the easiest ecosystem in aggregate, with the partial-carry trap as its
principal signal. Third, SemVerBench is a single snapshot tied to each ecosystem's
official resolver library, so resolver-/version-specific policy nuances (we filter
prereleases for PEP~440) are out of scope. Fourth, the main evaluation is
three-run (item-level Wilson CIs plus McNemar), whereas the mitigation conditions
(E1/E2/E3, against the three-run E0 baseline) and the false-premise study are each
single-run; we therefore read their causal interpretation as directional rather
than variance-controlled, and multi-run replication is future work. Fifth,
\emph{activation versus in-context teaching}: because both E1 and the TRUE hint make
the rule available in context, we cannot fully separate ``activating latent
knowledge'' from ``supplying missing knowledge''; the near-ceiling basic buckets and
the sufficiency of a light hint make the activation reading more parsimonious but not
decisive, and probing whether a model can state the rule unprompted would better
separate the two---future work. Sixth, \emph{per-bucket sample size}: some
main-benchmark buckets are small (the PEP~440 prefix-match bucket has $n=7$), so
bucket-level numbers are indicative; we address the most consequential case with the
67-item focused validation (Finding~3, Table~\ref{tab:prefix}) and otherwise rely on
three-run stability and cross-ecosystem consistency. Seventh,
\emph{external validity}: SemVerBench measures version-membership decisions in
isolation, and we do not directly measure how an in-head error propagates through
a coding agent into a real failure. The inference chain---an erroneous membership
verdict yielding an unsafe range or a missed patch, and thence a supply-chain
incident---is plausible and motivated by empirical evidence that breaking changes
do manifest in dependent packages~\cite{breakingchanges}, but that end-to-end link
is not established by our experiments. Relatedly, the three LLM proposers were
prompted for \emph{challenging} items, so SemVerBench deliberately targets
difficult/adversarial cases rather than the natural distribution of real-world
version-resolution queries; a repository-derived, real-workload test set is future
work. Eighth, \emph{scale and prompt
robustness}: the benchmark is 240 items (we prioritized machine-verified
correctness and mechanism coverage over raw scale). On prompt phrasing, we
additionally evaluated all 240 items under two alternative phrasings differing in
wording and structure;
per-model accuracy shifts by at most 4.2 points across the three phrasings
(GPT-5.1 by only 0.8), so the findings are not artifacts of prompt wording. Ninth, \emph{training contamination}:
the official test suites are public code likely present in pretraining corpora, so
the 92.8--99.4\% official-suite accuracies (Table~\ref{tab:coi}) may partly
reflect memorization rather than generalization. This does not affect the main
findings---difficulty comes from the adversarial LLM-proposed items, not the
official suites---but the official-suite numbers should be read as an easy anchor,
not as evidence of genuine generalization.

\noindent\textbf{Conclusion.} On a machine-checkable task with a unique answer,
frontier LLMs misapply version-constraint rules whose basic forms they handle
perfectly. The failures are systematic and predictable, not diffuse: a universal
partial-comparator carry trap drags every model to $\approx$60\% on Cargo, PEP~440
prefix corner cases (zero-pad and post-release) collapse GPT-5.1 while Claude is
near-perfect, ordered-comparison exclusion cracks even strong models, and Claude
leads significantly. As the more parsimonious of two readings we cannot fully
separate, the diagnosis is an activation/application gap rather than a knowledge
gap (Section~\ref{sec:mitig}).
Since a free, 100\%-correct resolver exists and tool delegation reaches
$\approx$100\%, the recommendation is unambiguous: \emph{coding agents must not
resolve version constraints in-head---they should delegate to the resolver.}

\section*{GenAI Usage Disclosure}
LLMs play three roles in this work. \textbf{(i) Subjects:} the six-model panel
(Claude-Opus, Claude-Sonnet, GPT-5.1, GPT-4.1, GPT-4o, Gemini-2.5-Pro) is the
object of study. \textbf{(ii) Item proposers:} three frontier LLMs (Claude, GPT,
Gemini) proposed part of the main 240 items from an identical difficulty-oriented
prompt (their suggested answers discarded; every item labeled by the oracle); we
record each proposer and analyze self-favoritism (Section~\ref{sec:coi}), and the
67-item focused-validation set (Finding~3) is author-constructed. \textbf{(iii) Assistance:} the authors used LLM-based
assistants to help with coding and to prepare an initial draft; all research
questions, design, analysis, verification, and conclusions are the authors' own.

\section*{Artifact Availability}
The benchmark data and the two-implementation ground-truth oracle (for
independent label verification) are available at:
\url{https://github.com/qibaic/semverbench}.

\end{document}